\documentclass[conference]{IEEEtran}
\IEEEoverridecommandlockouts

\usepackage{amsmath,amssymb,amsfonts}
\usepackage{algorithmic}
\usepackage{graphicx}
\usepackage{textcomp}
\usepackage{xcolor}
\usepackage{hyperref}
\usepackage{tikz}
\usepackage{pgfplots}
\usepackage{booktabs}
\usepackage{float}
\usetikzlibrary{arrows.meta}

\pgfplotsset{compat=1.18}

\usepackage[backend=biber,style=ieee]{biblatex}
\def\BibTeX{{\rm B\kern-.05em{\sc i\kern-.025em b}\kern-.08em
    T\kern-.1667em\lower.7ex\hbox{E}\kern-.125emX}}
\begin{document}

\title{\MakeUppercase{Residual Connections and the Causal Shift: Uncovering a Structural Misalignment in Transformers}}

\author{\IEEEauthorblockN{Jonathan Lys, Vincent Gripon,\\Bastien Pasdeloup, Axel Marmoret}
\\
\IEEEauthorblockA{IMT Atlantique, Lab-STICC,\\ UMR CNRS 6285, Brest, France \\
\href{mailto:name.surname@imt-atlantique.fr}{\texttt{name.surname@imt-atlantique.fr}}}
\and
\IEEEauthorblockN{Lukas Mauch, Fabien Cardinaux,\\Ghouthi Boukli Hacene}
\\
\IEEEauthorblockA{Sony Europe Ltd. \\ 
Stuttgart Technology Center, EUREC, Germany
\\
\href{mailto:Name.Surname@sony.com}{\texttt{Name.Surname@sony.com}}}
}

\maketitle

\begin{abstract}
Large Language Models (LLMs) are trained with next-token prediction, implemented in autoregressive Transformers via causal masking for parallelism. This creates a subtle misalignment: residual connections tie activations to the current token, while supervision targets the next token, potentially propagating mismatched information if the current token is not the most informative for prediction.
In this work, we empirically localize this input–output alignment shift in pretrained LLMs, using decoding trajectories over tied embedding spaces and similarity-based metrics. Our experiments\footnote{\scriptsize\url{https://github.com/jonathanlys01/causal-shift}} reveal that the hidden token representations switch from input alignment to output alignment deep within the network. Motivated by this observation, we propose a lightweight residual-path mitigation based on residual attenuation, implemented either as a fixed-layer intervention or as a learnable gating mechanism. Experiments on multiple benchmarks show that these strategies alleviate the representation misalignment and yield improvements, providing an efficient and general architectural enhancement for autoregressive Transformers.

\end{abstract}

\begin{IEEEkeywords}
LLM, Autoregressive, Causal Masking, Transformers
\end{IEEEkeywords}

\section{Introduction}\label{sec:intro}

The success of Transformer-based~\cite{vaswani2017attention} Large Language Models (LLMs) is rooted in architectural design that simultaneously prioritizes training stability and massive computational throughput. Transformer architectures rely on residual connections, originally introduced to stabilize optimization and facilitate gradient flow~\cite{xiong2020on}, that propagate representations across depth. In addition, Transformer-based LLMs trained via next-token prediction require parallelism for high hardware utilization~\cite{shoeybi2019megatron, narayanan2021efficient}. Decoder-only architectures like GPT~\cite{radford2019language} and LLaMA~\cite{grattafiori2024llama} suit this regime well: causal masking enables parallel processing where position $i$ attends only to tokens~$t_{\le i}$ to predict~$t_{i+1}$. This parallel loss computation~\cite{vaswani2017attention} significantly enhances training efficiency and scalability. However, this parallel computation also introduces a systematic one-token offset between inputs and supervision: the hidden state at position $i$ is initialized from the embedding of token $t_i$, and then optimized to predict $t_{i+1}$.

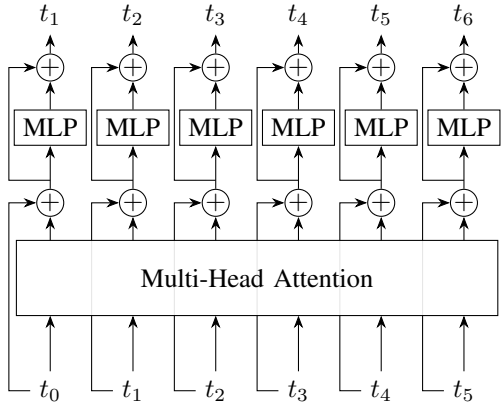
\begin{figure}[t]
    \centering
    \begin{tikzpicture}
        \node(0) at (0,0) {$t_0$};
        \node(1) at (1.1,0) {$t_1$};
        \node(2) at (2.2,0) {$t_2$};
        \node(3) at (3.3,0) {$t_3$};
        \node(4) at (4.4,0) {$t_4$};
        \node(5) at (5.5,0) {$t_5$};
        \node[draw, circle, inner sep = 0pt](00) at (0,2.5) {$+$};
        \node[draw, circle, inner sep = 0pt](11) at (1.1,2.5) {$+$};
        \node[draw, circle, inner sep = 0pt](22) at (2.2,2.5) {$+$};
        \node[draw, circle, inner sep = 0pt](33) at (3.3,2.5) {$+$};
        \node[draw, circle, inner sep = 0pt](44) at (4.4,2.5) {$+$};
        \node[draw, circle, inner sep = 0pt](55) at (5.5,2.5) {$+$};
        \path[-Stealth]
        (0) edge (0,1)
        (1) edge (1.1,1)
        (2) edge (2.2,1)
        (3) edge (3.3,1)
        (4) edge (4.4,1)
        (5) edge (5.5,1);
        \draw[-Stealth]
        (0) -- +(-0.55,0) -- +(-0.55,2.5) -- (00);
        \draw[-Stealth]
        (1) -- +(-0.55,0) -- +(-0.55,2.5) -- (11);
        \draw[-Stealth]
        (2) -- +(-0.55,0) -- +(-0.55,2.5) -- (22);
        \draw[-Stealth]
        (3) -- +(-0.55,0) -- +(-0.55,2.5) -- (33);
        \draw[-Stealth]
        (4) -- +(-0.55,0) -- +(-0.55,2.5) -- (44);
        \draw[-Stealth]
        (5) -- +(-0.55,0) -- +(-0.55,2.5) -- (55);
        \node[draw, rectangle, minimum width=6.4cm, minimum height=1cm, fill=white, fill opacity=0.9, text opacity=1](MHA) at (2.75,1.5) {Multi-Head Attention};
        \path[-Stealth]
        (0,2) edge (00)
        (1.1,2) edge (11)
        (2.2,2) edge (22)
        (3.3,2) edge (33)
        (4.4,2) edge (44)
        (5.5,2) edge (55);
        \node[draw, rectangle](MLP0) at (0,3.5) {MLP};
        \node[draw, rectangle](MLP1) at (1.1,3.5) {MLP};
        \node[draw, rectangle](MLP2) at (2.2,3.5) {MLP};
        \node[draw, rectangle](MLP3) at (3.3,3.5) {MLP};
        \node[draw, rectangle](MLP4) at (4.4,3.5) {MLP};
        \node[draw, rectangle](MLP5) at (5.5,3.5) {MLP};
        \path[-Stealth]
        (00) edge (MLP0)
        (11) edge (MLP1)
        (22) edge (MLP2)
        (33) edge (MLP3)
        (44) edge (MLP4)
        (55) edge (MLP5);
        \node[draw, circle, inner sep = 0pt](000) at (0,4.3) {$+$};
        \node[draw, circle, inner sep = 0pt](111) at (1.1,4.3) {$+$};
        \node[draw, circle, inner sep = 0pt](222) at (2.2,4.3) {$+$};
        \node[draw, circle, inner sep = 0pt](333) at (3.3,4.3) {$+$};
        \node[draw, circle, inner sep = 0pt](444) at (4.4,4.3) {$+$};
        \node[draw, circle, inner sep = 0pt](555) at (5.5,4.3) {$+$};

        \draw[-Stealth]
        (0,2.8) -- +(-0.55,0) -- +(-0.55,1.5) -- (000);
        \draw[-Stealth]
        (1.1,2.8) -- +(-0.55,0) -- +(-0.55,1.5) -- (111);
        \draw[-Stealth]
        (2.2,2.8) -- +(-0.55,0) -- +(-0.55,1.5) -- (222);
        \draw[-Stealth]
        (3.3,2.8) -- +(-0.55,0) -- +(-0.55,1.5) -- (333);
        \draw[-Stealth]
        (4.4,2.8) -- +(-0.55,0) -- +(-0.55,1.5) -- (444);
        \draw[-Stealth]
        (5.5,2.8) -- +(-0.55,0) -- +(-0.55,1.5) -- (555);
        \path[-Stealth]
        (MLP0) edge (000)
        (MLP1) edge (111)
        (MLP2) edge (222)
        (MLP3) edge (333)
        (MLP4) edge (444)
        (MLP5) edge (555);
        
        \node(o0) at (0,5) {$t_1$};
        \node(o1) at (1.1,5) {$t_2$};
        \node(o2) at (2.2,5) {$t_3$};
        \node(o3) at (3.3,5) {$t_4$};
        \node(o4) at (4.4,5) {$t_5$};
        \node(o5) at (5.5,5) {$t_6$};
        \path[-Stealth]
        (000) edge (o0)
        (111) edge (o1)
        (222) edge (o2)
        (333) edge (o3)
        (444) edge (o4)
        (555) edge (o5);
        
    \end{tikzpicture}
    \caption{Schematic illustration of the token misalignment between input and output in Transformer architectures (here with a single layer). Token $t_i$ in the input is directly connected to token $t_{i+1}$ in the output through the residual connections.}
    \label{fig:illustration}
\end{figure}

We conjecture that this offset creates a fundamental structural tension we characterize as \emph{input--output leakage}. On one hand, the residual connections act as a conservative bias, persistently propagating $t_{i}$ in the hidden state. On the other hand, the optimization process may drive the hidden state apart from $t_{i}$ to integrate context and yield accurate $t_{i+1}$, as show in Figure~\ref{fig:illustration}. Indeed, relying on $t_i$ to predict $t_{i+1}$ is a useful prior for local patterns, but it may be suboptimal when the prediction depends on long-range dependencies. Mechanistic interpretability research suggests that specific circuits, such as induction heads~\cite{olsson2022context}, override local context by copying information from earlier in the sequence. Consequently, as network depth increases, input-token--anchored components of the residual stream can interfere with representations that must ultimately become prediction-oriented. 

Thus, for accurate next-token prediction, representations must transition across depth from predominantly input-token-aligned to predominantly prediction-token-aligned. Characterizing where and how \textit{token alignment} transitions may be used to efficiently enhance next-token prediction. It also provides a simple interpretive axis for depth in LLMs through the alignment with both the input and output.

In this paper, we investigate this input-output leakage in two stages. First, we empirically characterize the depth at which internal representations transition from being primarily anchored to the input token to becoming aligned with the prediction target. Second, building on this observation, we introduce a lightweight modification of the residual pathway that selectively attenuates residual contributions through fixed or learned gating mechanisms. 

We therefore address two key questions:
\begin{enumerate}
\item At which layer(s) does the alignment shift occur, i.e., when does the model transition from input-anchored to prediction-oriented representations?
\item Given this transition point, how can residual pathways be adjusted to limit input-output leakage while preserving the stability benefits of residual connections?
\end{enumerate}

\section{Related Work}

Neural sequence modeling was originally tackled with Recurrent Neural Networks and their gradient stable variants Long Short-Term Memory networks~\cite{hochreiter1997long} and Gated Recurrent Units~\cite{cho2014learning}, which modeled temporal dependencies by maintaining a hidden state across time steps. In their formulation, the computation is inherently sequential, posing practical issues when scaling the model. The Transformer architecture~\cite{vaswani2017attention} introduced attention-based parallel sequence processing, greatly improving performance and scalability. Decoder-only variants, which use causal masking to enforce left-to-right dependencies, underpin modern LLMs such as Llama~\cite{grattafiori2024llama} and GPT~\cite{radford2018improving}.

Mechanistic interpretability (MI)~\cite{saphra2024mechanistic} has advanced our understanding of Transformer dynamics by analyzing how representations evolve across depth, e.g., by tracking hidden-state trajectories and decomposing the residual stream into component-wise contributions. Probing approaches such as the logit lens and its refined variants~\cite{nostalgebraist2020logitlens, belrose2023eliciting} map intermediate states into vocabulary space and show that prediction-relevant information emerges progressively across layers. Complementarily, authors in~\cite{csordas2025language} decompose layer and submodule contributions to token logits, and measure updates relative to the preceding residual state, finding that early layers induce large changes while later layers mostly refine. This staged behavior is consistent with the transition we identify in Sec.~\ref{sec:format}. Independent evidence on functional staging~\cite{lad2024remarkable} similarly suggests that earlier layers aggregate context whereas later layers primarily refine representations.

Our work is complementary to these analyses. Rather than focusing on local, layer-to-layer updates or intermediate prediction elicitation per se, we measure \emph{token alignment} relative to two global reference points, the input token embedding and the final prediction target, to explicitly characterize the transition from input-anchored to prediction-oriented representations, and relate it to persistent residual propagation under next-token supervision.

\section{Uncovering the shift location in pretrained models}
\label{sec:format}

Because the input and target sequences are offset by one position, a pretrained autoregressive Transformer must, at some depth, transition from representations primarily anchored to the current input token to representations oriented toward the next-token target, even if this transition is gradual and only loosely defined. In this section, we aim to localize this ``shift'' across layers through early layer decoding.

Many pretrained language models employ tied embeddings, where the same token-to-vector mapping (the dictionary) is shared between the embedding and unembedding layers. This means that the representation of a token at any depth can be ``decoded'' back to its nearest token in the dictionary, denoted as the \textit{logit lens}~\cite{nostalgebraist2020logitlens}. Consequently, tracking the evolution of the representation of a token in the depth of the model can be seen as following a trajectory through the Voronoi cells induced by this shared token dictionary. Note that this is not a common practice for large language models above 70B parameters (\cite{grattafiori2024llama, yang2025qwen3technicalreport, deepseekai2025deepseekv3technicalreport}).

Following the logit lens methodology, Table~\ref{tab:decoding} reports the tokens decoded from each intermediate representation of the Gemma-2-2B pretrained model~\cite{team2024gemma}, from the input embedding layer through to the final output layer, for a representative text sequence. More specifically, we highlight layers where the decoded token coincides with either the input token or the prediction target.

The results reveal three distinct regimes:
\begin{enumerate}
\item Input layers, where decoded tokens match the input sequence;
\item Intermediate layers, where decoding produces no meaningful correspondence;
\item Output layers, where decoded tokens match the shifted input sequence or semantically similar variants.
\end{enumerate}

\begin{table*}[t]
\renewcommand{\arraystretch}{1.2}

    \caption{Decoded hidden states of Gemma-2-2B on a Wikipedia article about the Fourier transform. Layers are grouped by similar decodings: early layers (0--6) align with the input (green), while late layers (21--26) align with the one-token-shifted targets (red). For instance, token \#107 consistently decodes to ``musical'' across layers 0--6, whereas token \#109 alternates between ``into'' and \texttt{<bos>} within the same range.}

    \centering
    \footnotesize
    \setlength{\tabcolsep}{2.5pt}
    \begin{tabular}{c | c c c c c c c c c}
    \toprule
    Layer & token \#107 & token \#108 & token \#109 & token \#110 & token \#111 & token \#112 & token \#113 \\
    \midrule
  Input & musical & chord & into & the & intensities & of & its \\ \midrule
  
  0 - 6 & \textcolor{olive}{musical}    &  \textcolor{olive}{chord}      &  \textcolor{olive}{into}, $<$ bos $>$       &  \textcolor{olive}{the}        &  \textcolor{olive}{intensities}, myself     &  \textcolor{olive}{of},  $<$ bos $>$        &  \textcolor{olive}{its} \\
  7 - 9 & \textcolor{olive}{musical}    &  \textcolor{olive}{chord}                 & AnchorStyles, These   &  \textcolor{olive}{the}        &  itself     &  lenker     &  \textcolor{olive}{its}       \\
 10     & \textcolor{olive}{musical}    &  chords     &  itself     &  \textcolor{olive}{the}        &  itself     & ReusableCell &  {\#\#\#\#}   \\
 11 & \textcolor{olive}{musical}    &  itself     & AddTagHelper &  various    &  itself     &  BoxDecoration &  {\#\#\#\#}   \\
 12 & \textcolor{olive}{musical}    &  itself     &  lenker     & WithMany    &  itself     &  lenker     & WebVitals   \\
 13 & \textcolor{olive}{musical}    &  itself     &  lenker     &  transfieras &  itself     &  lenker     &  {\#\#\#\#}  \\
 14 - 17    & \textcolor{olive}{musical}   &  itself     &  {Thefe}, lenker        & SequentialGroup, various, individual  &  itself     &  lenker, various     &  {\#\#\#\#}, {feveral} \\
 18 - 20    & \textcolor{olive}{musical}    &  \textcolor{red}{into}       &  individual, separate      &  individual &  itself, \textcolor{red}{of}        &  various    &  \textcolor{red}{constituent} \\
 21 - 26 & instrument &  \textcolor{red}{into}       &  its        &  various, frequency, frequencies     &  \textcolor{red}{of}         &  various, \textcolor{red}{its}, the   &  \textcolor{red}{constituent}, component\\
\bottomrule
    \end{tabular}
    \label{tab:decoding}
\end{table*}

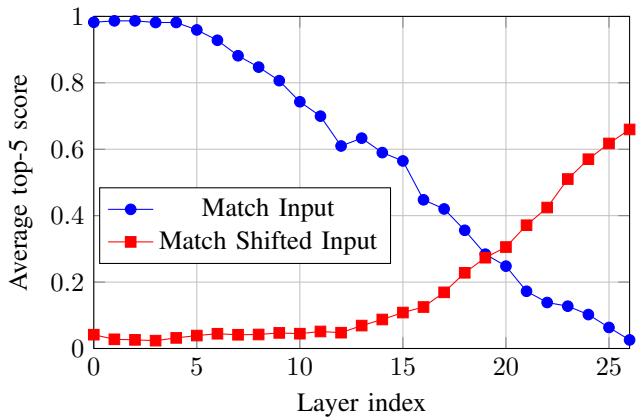
\begin{figure}[t]
    \centering
    \begin{tikzpicture}
\begin{axis}[
    width=.48\textwidth,
    height=6cm,
    xlabel={Layer index},
    ylabel={Average top-5 score},
    xmin=0, xmax=26,
    ymin=0, ymax=1,
    legend style={at={(0.01,0.37)},anchor=west},
    grid=both,
    grid style={line width=.1pt, draw=gray!20},
    major grid style={line width=.2pt,draw=gray!50},
]
\addplot[
    color=blue,
    mark=*,
] coordinates {
(0,0.9823389053344727) (1,0.9865767955780029) (2,0.9865767955780029) (3,0.9816293120384216) (4,0.9816490411758423) (5,0.959296703338623) (6,0.9281532764434814) (7,0.8815169334411621) (8,0.847278892993927) (9,0.8064573407173157) (10,0.7428892254829407) (11,0.699603796005249) (12,0.6097608804702759) (13,0.6333747506141663) (14,0.5898330211639404) (15,0.5647408962249756) (16,0.44771647453308105) (17,0.42018017172813416) (18,0.35594189167022705) (19,0.2840360403060913) (20,0.24798454344272614) (21,0.17233358323574066) (22,0.13845032453536987) (23,0.12743185460567474) (24,0.10218200832605362) (25,0.06333155930042267) (26,0.02582145668566227) };

\addlegendentry{Match Input}
\addplot[
    color=red,
    mark=square*,
] coordinates {
(0,0.0416099987924099) (1,0.027654584497213364) (2,0.025998856872320175) (3,0.0236138217151165) (4,0.03191216662526131) (5,0.03894900903105736) (6,0.04444838687777519) (7,0.041373465210199356) (8,0.042142193764448166) (9,0.0469713993370533) (10,0.04456665366888046) (11,0.051268406212329865) (12,0.04762186482548714) (13,0.06896891444921494) (14,0.08735930919647217) (15,0.10821358859539032) (16,0.12490883469581604) (17,0.16908127069473267) (18,0.22795812785625458) (19,0.2734117805957794) (20,0.3054816424846649) (21,0.37115880846977234) (22,0.42441803216934204) (23,0.5102201700210571) (24,0.5699643492698669) (25,0.6173693537712097) (26,0.6596) };
\addlegendentry{Match Shifted Input}

\end{axis}
\end{tikzpicture}
    \caption{Average top-5 match rate between the decoded hidden states and either the input sequence (blue) or the shifted input sequence (red) as a function of layer depth in the Gemma-2-2B model. Results are averaged over 1,000 sequences from the Wikitext dataset. The shift from input to output indexing occurs late in the architecture.}
    \label{fig:shiftstatsvoronoi}
\end{figure}

To quantify this transition, we measure at each layer how often the hidden state decodes to the input token ($t_i$) or to the shifted token ($t_{i+1}$), i.e., the prediction target. We evaluate this on 1,000 randomly sampled sequences from Wikitext~\cite{merity2016pointer}, using a top-5 match criterion to account for both exact and near matches.
This provides a systematic, layer-wise measure of token alignment rather than relying on isolated examples. As shown in Figure~\ref{fig:shiftstatsvoronoi}, the shift from input-aligned to target-aligned decoding occurs relatively deep in the network, indicating that a substantial fraction of the forward pass remains indexed to the input token before becoming prediction-oriented.

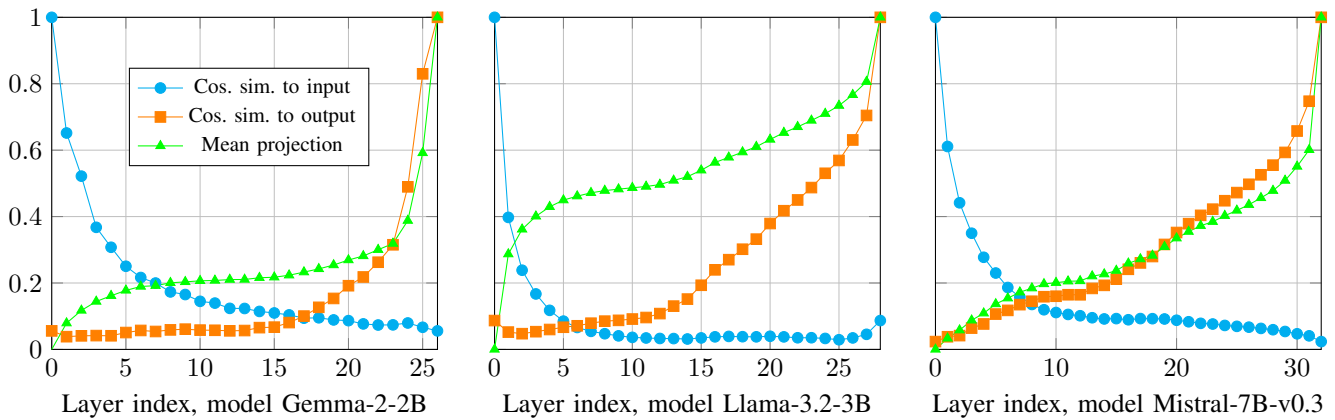
\begin{figure*}[ht]
    \centering
\begin{tabular}{ccc}
\begin{tikzpicture}
\begin{axis}[
    width=.37\textwidth,
    height=6cm,
    xlabel={Layer index, model Gemma-2-2B},
    xmin=0, xmax=26,
    ymin=0, ymax=1,
    legend style={at={(0.2,0.7)},anchor=west},
    grid=both,
    grid style={line width=.1pt, draw=gray!20},
    major grid style={line width=.2pt,draw=gray!50},
]
\addplot[
    color=cyan,
    mark=*,
] coordinates {
(0,1.0) (1,0.6517082452774048) (2,0.5217319130897522) (3,0.36764609813690186) (4,0.30759912729263306) (5,0.25038567185401917) (6,0.21651089191436768) (7,0.19984640181064606) (8,0.17315177619457245) (9,0.16524574160575867) (10,0.1448030024766922) (11,0.139469176530838) (12,0.12361791729927063) (13,0.12326715141534805) (14,0.11438266932964325) (15,0.10976000875234604) (16,0.10395485907793045) (17,0.09413008391857147) (18,0.09523984044790268) (19,0.0888146311044693) (20,0.08665546029806137) (21,0.07682777941226959) (22,0.07346347719430923) (23,0.07392434030771255) (24,0.07959494739770889) (25,0.06665419042110443) (26,0.05580070987343788) };

\addlegendentry{\footnotesize{Cos. sim. to input}}
\addplot[
    color=orange,
    mark=square*,
] coordinates {
(0,0.05580070987343788) (1,0.03834975138306618) (2,0.0409291572868824) (3,0.041613660752773285) (4,0.041105955839157104) (5,0.050805576145648956) (6,0.057098422199487686) (7,0.053987689316272736) (8,0.05878438428044319) (9,0.06095336750149727) (10,0.05810876563191414) (11,0.05763794854283333) (12,0.05599873140454292) (13,0.05684998258948326) (14,0.0656028464436531) (15,0.06701824069023132) (16,0.08081627637147903) (17,0.10030222684144974) (18,0.12673319876194) (19,0.153507262468338) (20,0.19183248281478882) (21,0.21811014413833618) (22,0.26254865527153015) (23,0.31509286165237427) (24,0.48959872126579285) (25,0.8299399018287659) (26,1.0)};
\addlegendentry{\footnotesize{Cos. sim. to output}}
\addplot[
    color=green,
    mark=triangle*,
] coordinates {
(0,-7.807998190401122e-05) (1,0.07916699349880219) (2,0.11736905574798584) (3,0.14432880282402039) (4,0.1615494191646576) (5,0.1777341067790985) (6,0.18951867520809174) (7,0.19196568429470062) (8,0.20054244995117188) (9,0.20336538553237915) (10,0.20661227405071259) (11,0.20760007202625275) (12,0.2103118598461151) (13,0.21057480573654175) (14,0.21579821407794952) (15,0.21734537184238434) (16,0.22339776158332825) (17,0.23276181519031525) (18,0.24246099591255188) (19,0.25412696599960327) (20,0.26921147108078003) (21,0.281467080116272) (22,0.30016472935676575) (23,0.31879308819770813) (24,0.38793906569480896) (25,0.5917440056800842) (26,1.0)};
\addlegendentry{\footnotesize{Mean projection}}

\end{axis}
\end{tikzpicture}
&
\begin{tikzpicture}
\begin{axis}[
    width=.37\textwidth,
    height=6cm,
    xlabel={Layer index, model Llama-3.2-3B},
    xmin=0, xmax=28,
    ymin=0, ymax=1,
    yticklabel=\empty,
    legend pos=north east,
    grid=both,
    grid style={line width=.1pt, draw=gray!20},
    major grid style={line width=.2pt,draw=gray!50},
]
\addplot[
    color=cyan,
    mark=*,
] coordinates {
(0,1.0) (1,0.3975714445114136) (2,0.2385125309228897) (3,0.1670522391796112) (4,0.11764229089021683) (5,0.08515814691781998) (6,0.0660441666841507) (7,0.054379578679800034) (8,0.0472344234585762) (9,0.04084497690200806) (10,0.03628550097346306) (11,0.03443380445241928) (12,0.03294624760746956) (13,0.03263916075229645) (14,0.031377971172332764) (15,0.03459196165204048) (16,0.03787391632795334) (17,0.039152346551418304) (18,0.03831838071346283) (19,0.03790116682648659) (20,0.03959525004029274) (21,0.03777159005403519) (22,0.035083334892988205) (23,0.03519438952207565) (24,0.03305262699723244) (25,0.029122285544872284) (26,0.03461340069770813) (27,0.04544133320450783) (28,0.08690033107995987) };

\addplot[
    color=orange,
    mark=square*,
] coordinates {
(0,0.08690033107995987) (1,0.05215831473469734) (2,0.0473833791911602) (3,0.05339328572154045) (4,0.059972506016492844) (5,0.06677158921957016) (6,0.07117629051208496) (7,0.07892213016748428) (8,0.0850745365023613) (9,0.0875338539481163) (10,0.09141489863395691) (11,0.09606269001960754) (12,0.10788977891206741) (13,0.1303478330373764) (14,0.15152591466903687) (15,0.19318614900112152) (16,0.2396707683801651) (17,0.2702041566371918) (18,0.3016822338104248) (19,0.3320119082927704) (20,0.37908411026000977) (21,0.41762638092041016) (22,0.45002439618110657) (23,0.48733654618263245) (24,0.53037428855896) (25,0.5689908862113953) (26,0.6306650638580322) (27,0.7044968008995056) (28,1.0)};
\addplot[
    color=green,
    mark=triangle*,
] coordinates {
(0,0.0) (1,0.2870468497276306) (2,0.36156928539276123) (3,0.40083232522010803) (4,0.4294430911540985) (5,0.44959816336631775) (6,0.4615141451358795) (7,0.47153738141059875) (8,0.47821319103240967) (9,0.48261547088623047) (10,0.4867197573184967) (11,0.49054884910583496) (12,0.49640175700187683) (13,0.5088047981262207) (14,0.5201287269592285) (15,0.5395205616950989) (16,0.562867283821106) (17,0.5783206224441528) (18,0.5941265821456909) (19,0.6102876663208008) (20,0.6321915984153748) (21,0.6526291966438293) (22,0.6701198220252991) (23,0.6894481182098389) (24,0.7098897099494934) (25,0.7341940999031067) (26,0.7666110992431641) (27,0.8061110973358154) (28,1.0)};

\end{axis}
\end{tikzpicture}
&
\begin{tikzpicture}
\begin{axis}[
    width=.37\textwidth,
    height=6cm,
    xlabel={Layer index, model Mistral-7B-v0.3},
    xmin=0, xmax=32,
    ymin=0, ymax=1,
    yticklabel=\empty,
    legend pos=north east,
    grid=both,
    grid style={line width=.1pt, draw=gray!20},
    major grid style={line width=.2pt,draw=gray!50},
]
\addplot[
    color=cyan,
    mark=*,
] coordinates {
(0,1.0) (1,0.6110242009162903) (2,0.44150328636169434) (3,0.34987378120422363) (4,0.2774451673030853) (5,0.23001544177532196) (6,0.18645015358924866) (7,0.15739190578460693) (8,0.13532426953315735) (9,0.1194329485297203) (10,0.11105193942785263) (11,0.1053200215101242) (12,0.10135604441165924) (13,0.09537820518016815) (14,0.09177891165018082) (15,0.09275411814451218) (16,0.08932562172412872) (17,0.09284177422523499) (18,0.09254317730665207) (19,0.0916438177227974) (20,0.08781855553388596) (21,0.08328215777873993) (22,0.07853332161903381) (23,0.07647308707237244) (24,0.07221769541501999) (25,0.0694616436958313) (26,0.06699775904417038) (27,0.0632476732134819) (28,0.059294845908880234) (29,0.05399094149470329) (30,0.047035593539476395) (31,0.040885597467422485) (32,0.023547550663352013)
};

\addplot[
    color=orange,
    mark=square*,
] coordinates {
(0,0.023547550663352013) (1,0.038134895265102386) (2,0.04159460961818695) (3,0.06425744295120239) (4,0.07665399461984634) (5,0.10644368827342987) (6,0.11755319684743881) (7,0.13484416902065277) (8,0.1450463831424713) (9,0.15864308178424835) (10,0.1601400524377823) (11,0.16462786495685577) (12,0.1645413041114807) (13,0.1837085336446762) (14,0.19294849038124084) (15,0.21153783798217773) (16,0.24127095937728882) (17,0.2605099380016327) (18,0.27974942326545715) (19,0.3160820007324219) (20,0.352012038230896) (21,0.3786572813987732) (22,0.4036726653575897) (23,0.4222545027732849) (24,0.44772613048553467) (25,0.47218477725982666) (26,0.49724990129470825) (27,0.5258274078369141) (28,0.5547866821289062) (29,0.5931907892227173) (30,0.6576075553894043) (31,0.7473576068878174) (32,1.0)
};
\addplot[
    color=green,
    mark=triangle*,
] coordinates {
(0,0.0) (1,0.0333707295358181) (2,0.0579863004386425) (3,0.08768337219953537) (4,0.1089264526963234) (5,0.1367468684911728) (6,0.15420135855674744) (7,0.1721676141023636) (8,0.18445107340812683) (9,0.19707363843917847) (10,0.2008381187915802) (11,0.2052631974220276) (12,0.20612552762031555) (13,0.22060483694076538) (14,0.22668981552124023) (15,0.23728948831558228) (16,0.25923416018486023) (17,0.2721133232116699) (18,0.28283846378326416) (19,0.3089090585708618) (20,0.3349299728870392) (21,0.3547438979148865) (22,0.37238645553588867) (23,0.3845677673816681) (24,0.40257415175
43793) (25,0.4184570610523224) (26,0.43570971488952637) (27,0.4562768340110779) (28,0.47801387310028076) (29,0.5085054636001587) (30,0.5509401559829712) (31,0.6010472178459167) (32,1.0)
};

\end{axis}
\end{tikzpicture}
\end{tabular}
\caption{Continuous similarity measures between hidden states and their corresponding input and shifted input tokens, reported for Gemma-2-2B~\cite{team2024gemma}, Llama-3.2-3B~\cite{grattafiori2024llama} and Mistral-7B-v0.3~\cite{jiang2023mistral7b}. Metrics include cosine similarity to input and output token embeddings, and normalized projection along the input–output axis (0: input token, 1: output token). Results confirm the latent transition from input-based to output-based representations across different architectures.}
    \label{fig:softermeasures}
\end{figure*}

To mitigate thresholding effects inherent to top-5 accuracy, we additionally report in Figure~\ref{fig:softermeasures} continuous similarity measures. Concretely, we compute (i) the cosine similarity between each hidden state and the embeddings of the input and output tokens, and (ii) the mean projection onto the axis defined by the input-output token pair, normalized such that 0 corresponds to the input token and 1 to the output token.
We replicate this analysis on two additional widely used language models (Llama-3.2-3B~\cite{grattafiori2024llama} and Mistral-7B-v0.3~\cite{jiang2023mistral7b}). The results confirm that, in the Gemma-2-2B architecture, the transition occurs around layer 17, while other models exhibit different but qualitatively similar transition depths.

\section{Residual Path Decoupling}

In this section, we propose a simple mitigation for the input-output interference identified above by selectively attenuating the residual pathway beyond a chosen depth, thereby reducing the persistence of input-anchored components in the residual stream. All experiments follow the same training protocol: we train 150M-parameter GPT-2–style models on 10B tokens from the Fineweb-Edu dataset~\cite{penedo2024the}.

We introduce residual attenuation as:

\[x_{l+1}=\alpha x_l+F_l(\mathrm{LN}(x_l))\]

Where $\alpha$ is a scalar gating parameter. Setting $\alpha=0$ removes the skip connection but also eliminates the identity gradient path. Gradients must then flow entirely through $F_l$ (notably through attention and its softmax) where they are typically less stable than along the residual branch~\cite{xiong2020on}. Empirically, fully suppressing the residual connection degraded optimization, so we restrict $\alpha>0$ and focus on attenuation rather than removal.

Two main challenges arise:

\begin{itemize}
    \item As shown earlier, there is no sharply defined layer where the shift occurs, making the selection of a cutting point approximate.
    \item Even within a single sequence, different tokens may transition from input-based to output-based representations at different depths.
\end{itemize}

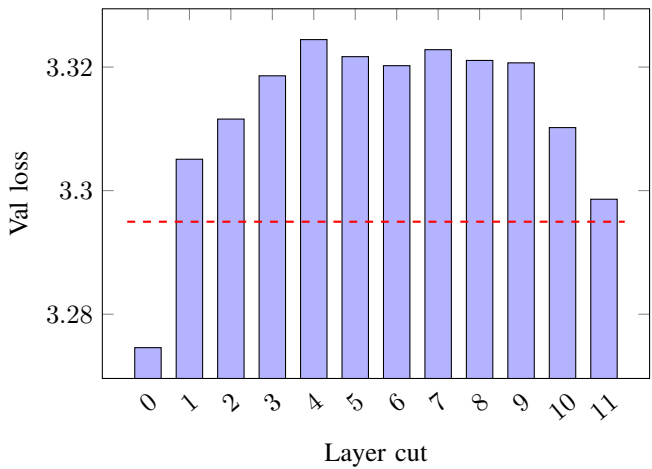
\begin{figure}[t!]
    \centering
\begin{tikzpicture}
\begin{axis}[
  width=\linewidth, 
  height=6.5cm,
  ylabel={Val loss},
  xlabel={Layer cut},
  xtick=data,
  xticklabel style={rotate=40},
  enlarge x limits=0.05,
  legend pos=north west,
]

\addplot[ybar,fill=blue!30,area legend]
  table[x expr=\coordindex,y index=1,col sep=comma,header=false,ignore chars={"}]{data/cut.csv};

\addplot[thick,dashed,red]
  coordinates {(-0.5,3.29497) (11.5,3.29497)};

\end{axis}
\end{tikzpicture}
    \caption{Impact of cutting the residual path at different layers in GPT2-0.1B, trained on 10B tokens from Fineweb~\cite{penedo2024the} on the validation loss. The baseline loss is represented with a dotted line.}
    \label{fig:residualcut}
\end{figure}

To evaluate fixed residual attenuation, we perform layer-wise ablations in which the residual branch is attenuated at a single layer at a time, covering all $12$ possible depths. For each configuration, we measure the validation loss and report the results in Figure~\ref{fig:residualcut}. Notably, we observe that attenuating only the first layer leads to a consistent improvement in validation performance.

 \begin{figure}[ht!]
     \begin{tikzpicture}
  \def\singlecolor{blue}
  \def\multicolor{red}
  
  \begin{axis}[
    height=6.5cm,
    width=\linewidth,
    xlabel={Training step},
    ylabel={Gate value},
        grid=both,
    grid style={line width=.1pt, draw=gray!20},
    major grid style={line width=.2pt,draw=gray!50},
    legend style={at={(0.7,0.7)}, anchor=north},
    ymin=0,
    ymax=1,
    xmin=0,
    xmax=200
  ]

    \addplot+[\singlecolor, thick]
      table[col sep=comma, x expr=\coordindex*27.5, y index=0]{data/single.csv};
    \addlegendentry{Last layer}

    \foreach \i in {0,...,9}{
      \addplot+[\multicolor, very thin, opacity=0.6, forget plot]
        table[col sep=comma, x expr=\coordindex*27.5, y index=\i]{data/multi.csv};
    }
    \addlegendimage{\multicolor, thick}
    \addlegendentry{Others}

  \end{axis}
\end{tikzpicture}
     \caption{Evolution of the learned probability distribution of cutting the residual path at a given layer, during training. At the beginning, all 12 layers have the same probability of cut, then the 11th layer attracts all the mass of the probability.}
     \label{fig:learnpath}
 \end{figure}
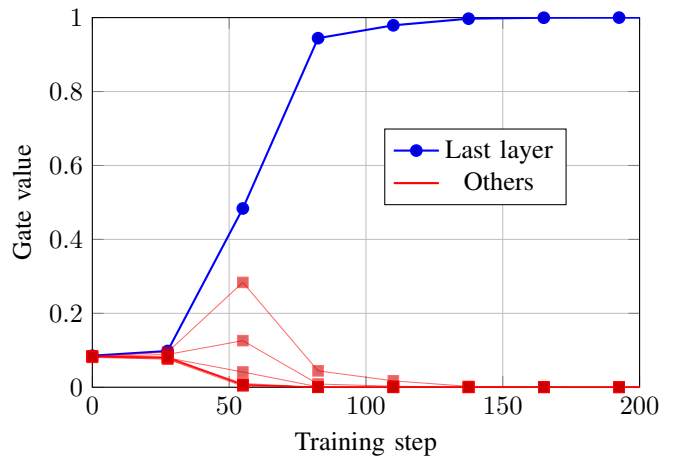

An alternative is to let the model learn where to attenuate the residual pathway. We introduce a mixture-of-depth–inspired gating mechanism~\cite{raposo2024mixture} that predicts, during training, where to downweight the residual branch. The gating distribution is initialized uniformly over layers and optimized jointly with the model, without enforcing a one-hot selection.

We observe (Figure~\ref{fig:learnpath}) that the learned distribution progressively concentrates on attenuating the final layer, corresponding to the second-best fixed choice in our earlier ablation. Despite this, the learned variant surpasses the performance obtained by statically cutting that layer from initialization, suggesting that adaptive attenuation benefits from smoother optimization dynamics.

Table~\ref{tab:finalresults} compares the baseline, fixed attenuation, and learnable gating across multiple benchmarks, including Wikitext~\cite{merity2016pointer}, LAMBADA~\cite{lambada}, and OpenWebText~\cite{Gokaslan2019OpenWeb}. Across all evaluated benchmarks, the learnable gating approach consistently outperforms fixed cuts and matches or surpasses baseline accuracy while reducing measured misalignment between input and output representations. The fixed-cut method offers modest improvements when the cut is placed late in the network but is highly sensitive to the chosen layer. In contrast, gating adapts automatically to each dataset and model depth, suggesting it is a robust and low-cost architectural enhancement for autoregressive LLMs. Overall, these results indicate that soft, learned residual attenuation constitutes a promising architectural refinement for autoregressive language models.

\begin{table}[t!]
    \caption{Comparison of the baseline, fixed attenuation, and learnable gating across multiple benchmarks: Wikitext~\cite{merity2016pointer}, LAMBADA~\cite{lambada}, OpenWebText~\cite{Gokaslan2019OpenWeb}, and Fineweb-Edu dataset~\cite{penedo2024the}. Bold scores are the best ones.}
\centering
    \begin{tabular}{cccc}
    \toprule
    benchmark & baseline & fixed-cut & gating\\
    \midrule

    Wikitext & 28.46&  28.62&  \textbf{28.75}\\ 
    LAMBADA & 32.99& 33.38& \textbf{33.52}\\ 
    OpenWebText &  34.74& 34.84&\textbf{35.50}\\ 
    Fineweb-Edu & 38.87&  38.98& \textbf{39.19}\\ 
    \bottomrule
    \end{tabular}
    \label{tab:finalresults}

\end{table}

\section{Conclusion}

In this work, we have shown that causal masking in autoregressive Transformers leads to a structural misalignment between input and output token representations. Our analysis across multiple LLMs revealed that the shift from input-based to output-based indexing occurs deep in the layers, but not uniformly across tokens. This observation motivated two intervention strategies: 1) cutting residual paths at fixed layers, and 2) introducing learnable residual gating. While fixed cuts can improve performance when applied late in the network, they are highly sensitive to the chosen depth. In contrast, residual gating allows the model to automatically attenuate misaligned residual contributions, and has shown to consistently achieve competitive or superior performance across benchmarks, while reducing misalignment metrics. These findings suggest that soft, learnable control over residual flow offers a promising direction for improving the representational integrity of causal LLMs without sacrificing efficiency.

\section{Acknowledgements}

This research has been funded, either in full or in part, by the French National Research Agency (ANR) under project ANR-24-CE23-7365. With a view to its publication in open access, the author has applied for an open access CC-BY licence for any manuscript accepted for publication (AAM) resulting from this submission. This work was granted access to the HPC resources of IDRIS under the allocation 2024-AD011015938 made by GENCI.

\printbibliography

\end{document}